\pdfoutput=1
\documentclass[11pt]{article}

\usepackage[preprint]{acl}

\usepackage{times}
\usepackage{latexsym}

\usepackage[T1]{fontenc}
\usepackage[utf8]{inputenc}

\usepackage{microtype}

\usepackage{inconsolata}

\usepackage{graphicx}

\definecolor{hycolor}{rgb}{0.7,0.7,0.3} % Yellow&Green

\title{SeMe: Training-Free Language Model Merging via Semantic Alignment}

\author{Jian Gu \\
  Monash University \\
  \texttt{jian.gu@monash.edu} \\\And
  Aldeida Aleti \\
  Monash University \\
  \texttt{aldeida.aleti@monash.edu} \\\AND
  Chunyang Chen \\
  Technical University of Munich \\
  \texttt{chun-yang.chen@tum.de} \\\And
  Hongyu Zhang \\
  Chongqing University \\
  \texttt{hyzhang@cqu.edu.cn} \\
}

\usepackage{mystyle}

\DeclareAcronym{api}{
	short = API,
	long = {Application Program Interface}
}

\DeclareAcronym{awgn}{
	short = AWGN,
	long = {additive white Gaussian noise}
}

\DeclareAcronym{vae}{
	short = VAE,
	long = {Variational AutoEncoder}
}

\DeclareAcronym{bert}{
	short = BERT,
	long = {Bidirectional Encoder Representations from Transformers}
}

\DeclareAcronym{roberta}{
	short = RoBERTa,
	long = {Robustly optimized BERT approach}
}

\DeclareAcronym{ast}{
	short = AST,
	long = {Abstract Syntax Tree}
}

\DeclareAcronym{bpe}{
	short = BPE,
	long = {Byte-Pair Encoding}
}

\DeclareAcronym{cfg}{
	short = CFG,
	long = {Control Flow Graph}
}

\DeclareAcronym{dcg}{
	short = DCG,
	long = {Discounted Cumulative Gain}
}

\DeclareAcronym{gpt}{
	short = GPT,
	long = {Generative Pretrained Transformer}
}

\DeclareAcronym{ir}{
	short = IR,
	long = {Information Retrieval}
}

\DeclareAcronym{lstm}{
	short = LSTM,
	long = {Long Short-Term Memory}
}

\DeclareAcronym{clm}{
	short = CLM,
	long = {Casual Language Modeling}
}

\DeclareAcronym{mlm}{
	short = MLM,
	long = {Masked Language Modeling}
}

\DeclareAcronym{mem}{
	short = MEM,
	long = {Multimodal Embedding Model}
}

\DeclareAcronym{cp}{
	short = CP,
	long = {Continuous Pretraining}
}

\DeclareAcronym{if}{
	short = IF,
	long = {Intermediate Finetuning}
}

\DeclareAcronym{mmpf}{
	short = MMPF,
	long = {Massive Multitask Pre-Finetuning}
}

\DeclareAcronym{aif}{
	short = AIF,
	long = {Adaptive Intermediate Finetuning}
}

\DeclareAcronym{mrr}{
	short = MRR,
	long = {Mean Reciprocal Rank}
}

\DeclareAcronym{ndcg}{
	short = NDCG,
	long = {Normalized Discounted Cumulative Gain}
}

\DeclareAcronym{nlp}{
	short = NLP,
	long = {Natural Language Processing}
}

\DeclareAcronym{nlp_pt}{
	short = NLP\textsubscript{PT},
	long = {Next Line Prediction}
}

\DeclareAcronym{nmt}{
	short = NMT,
	long = {Neural Machine Translation}
}

\DeclareAcronym{nsp}{
	short = NSP,
	long = {Next Sentence Prediction}
}

\DeclareAcronym{rnn}{
	short = RNN,
	long = {Recurrent Neural Network}
}

\DeclareAcronym{cnn}{
	short = CNN,
	long = {Convolutional Neural Network}
}

\DeclareAcronym{tf-idf}{
	short = tf-idf,
	long = {term frequency–-inverse document frequency}
}

\DeclareAcronym{anova}{
	short = ANOVA,
	long = {ANalysis Of VAriance}
}

\DeclareAcronym{da}{
	short = DA,
	long = {Domain-Adaptive}
}

\DeclareAcronym{ta}{
	short = TA,
	long = {Task-Adaptive}
}

\DeclareAcronym{ma}{
	short = MA,
	long = {Multiphase Adaptive}
}

\DeclareAcronym{ca}{
	short = CA,
	long = {Concept Annotation}
}

\DeclareAcronym{ce}{
	short = CE,
	long = {Concept Extrapolation}
}

\DeclareAcronym{ci}{
	short = CI,
	long = {Concept Interpolation}
}

\DeclareAcronym{gru}{
	short = GRU,
	long = {Gated Recurrent Unit}
}

\DeclareAcronym{sota}{
	short = SOTA,
	long = {state-of-the-art}
}

\DeclareAcronym{lcs}{
	short = LCS,
	long = {Longest Common Sequences}
}

\begin{document}
\maketitle
\begin{abstract}
% TODO update it
Despite the remarkable capabilities of Language Models (LMs) across diverse tasks, no single model consistently outperforms others, necessitating efficient methods to combine their strengths without expensive retraining. Existing model merging techniques—such as parameter averaging and task-guided fusion—often rely on data-dependent computations or fail to preserve internal knowledge, limiting their robustness and scalability.
We introduce SeMe (Semantic-based Merging), a novel, data-free, and training-free approach that leverages latent semantic alignment to merge LMs at a fine-grained, layer-wise level. Unlike prior work, SeMe not only preserves model behaviors but also explicitly stabilizes internal knowledge, addressing a critical gap in LM fusion. Through extensive experiments across diverse architectures and tasks, we demonstrate that SeMe outperforms existing methods in both performance and efficiency while eliminating reliance on external data. Our work establishes a new paradigm for knowledge-aware model merging and provides insights into the semantic structure of LMs, paving the way for more scalable and interpretable model composition.
\end{abstract}

\section{Introduction}
\label{sec:introduction}

Language Models (LMs) have demonstrated exceptional capabilities across a wide range of tasks. However, the diversity in their architectures, training data, and fine-tuning strategies has resulted in a growing zoo of specialized models, each excelling in different areas. Despite their strengths, no single model consistently outperforms others across all tasks. This motivates a central question: \emph{Can we combine the capabilities of multiple LMs into a single model without expensive retraining or access to original data?}

To address this, recent research has explored collaborative strategies among LMs. Among these, \emph{model merging}—which directly fuses the parameters of different models into a unified model—has emerged as a particularly attractive solution. Unlike output-level ensemble methods or pipeline-based cooperation frameworks, model merging aims to combine models in the parameter space to inherit their respective strengths while maintaining efficiency at inference.

Existing model merging approaches largely fall into two categories. The first involves naive or weighted averaging of model parameters, such as Model Soup~\cite{Wortsman2022ModelSoup} and Fisher merging~\cite{Matena2021Fisher}. These methods assume a common initialization and often yield promising results when the models are fine-tuned variants of the same backbone. The second category incorporates task semantics, using tools like task vectors~\cite{Ilharco2023TaskArithmetic} or conflict-aware weighting~\cite{Yadav2023TIES} to guide the merge process. However, most of these techniques rely on data sampling or additional training to compute merge coefficients, introducing additional computational costs, uncertainty, and potential bias.

In addition, in the prior work, LM merging focuses on perserve the prediction behaviors of source models. However, these methods ignore the knowledge of each model, and therefore, tend to cause potential risks.
Based on the recent work on the knowledge of LMs, more knowledge of LMs are separate from their behaviors, and ignoring the internal knowledge tend to cause side effects in various tasks~\emph{Orgad2024LLMsKM}.

In this work, we propose a new paradigm for model merging that eliminates the dependency on data and training. Leveraging semantic alignment across models—derived from their latent representations—we develop a data-free, computation-efficient merging method that is robust across tasks and fine-tuning regimes.
It is a fine-grained approach, not merely maintain the behaviors of LMs, but also stabilize the knowledge of LMs.
This \ul{se}mantic-based \ul{me}rging, shortened as \textbf{SeMe}, offers a scalable alternative to existing approaches and opens up new avenues for efficient model reuse in the LLM era.
% ...
The replication repository is attached as supplementary material.

Our contributions are as follows:

\smallskip
\noindent

\begin{itemize}
    \item We recognize the vector nature of semantics in LM latent space, and verify its correctness with empirical evidence. It may inspire future work on LM semantics;
    \item We implement a fine-grained (layer-wise) approach on LM merging, which eliminates the training needs and therefore reduced the computation costs;
    \item We propose aligning the internal knowledge, not merely external knowledge, as the solution to the research gap of LM fusion. We conduct extensive experiments with diverse models to support the claims.
\end{itemize}

\section{Background}
\label{sec:background}

Model merging aims to integrate multiple language models into a single, unified model by combining their parameters in the weight space. This process generally involves three key stages: aligning models within a common parameter space, estimating the importance of model-specific parameter differences, and aggregating these differences into a merged representation. We describe this as a three-step pipeline:

\paragraph{Select}
For each model, we compute a \textit{fusion vector}—the difference between the model and a shared pivot model. Across the set of fusion vectors, we identify high-variance components within each parameter matrix (e.g., attention or feedforward weights). These components are assumed to reflect salient, task-specific adaptations introduced during fine-tuning. Selection retains only the most informative elements (e.g., top-$\tau\%$ by variance), reducing noise and redundancy.

\paragraph{Compute}
Merge coefficients are calculated for each model’s selected components. A common approach is to assign weights in proportion to the squared magnitudes of the selected entries, such that models with stronger or more consistent updates contribute more to the merged result.

\paragraph{Erase}
To prevent conflicting parameter updates from causing destructive interference, entries with minority sign alignment (e.g., those that disagree in direction with most other models) are discarded. This step reduces parameter interference and improves merge stability.

The final merged model is obtained by adding the weighted, filtered fusion vectors back to the pivot model:
\begin{equation}
\theta_{\text{merge}} = \theta_{\text{pivot}} + \sum_{k=1}^{K} \eta_k \cdot \delta_k^\prime,
\end{equation}
where $\delta_k^\prime$ are the post-processed fusion vectors, and $\eta_k$ are the computed merge coefficients.

\section{Motivational Analysis}
\label{sec:preliminary}

% TODO background for modular LMs:
% \subsection{Configurable Foundation Models}
% The term, ``configurable language models'' refer to the modular design of language models, where reusable modules (of varying granularity and functionality) are denoted as functional components, derived from decomposing LLMs, terms as ``bricks'', including pretrained bricks of emergent abilities and customized bricks of special abilities~\citep{Xiao2024ConfigurableFM}.
% % 
% Modular language models aim to fuse multiple functional modules to fulfill complex instructions.
% The modular perspective of language models is promising in:
% (1) allows for selective inference to enhance efficiency;
% (2) allows for partial module maintenance to guarantee flexibility;
% (3) allows for detachable module assembly to realize continual learning.
% Besides, benefiting from the flexibility and scale effects, modular language models pave the way for powerful performance in addressing complex tasks across diverse environments.

\subsection{Preliminary: Vocabulary-Defined Semantics}

\begin{figure}[!htb]
    \centering
    \includegraphics[width=\linewidth]{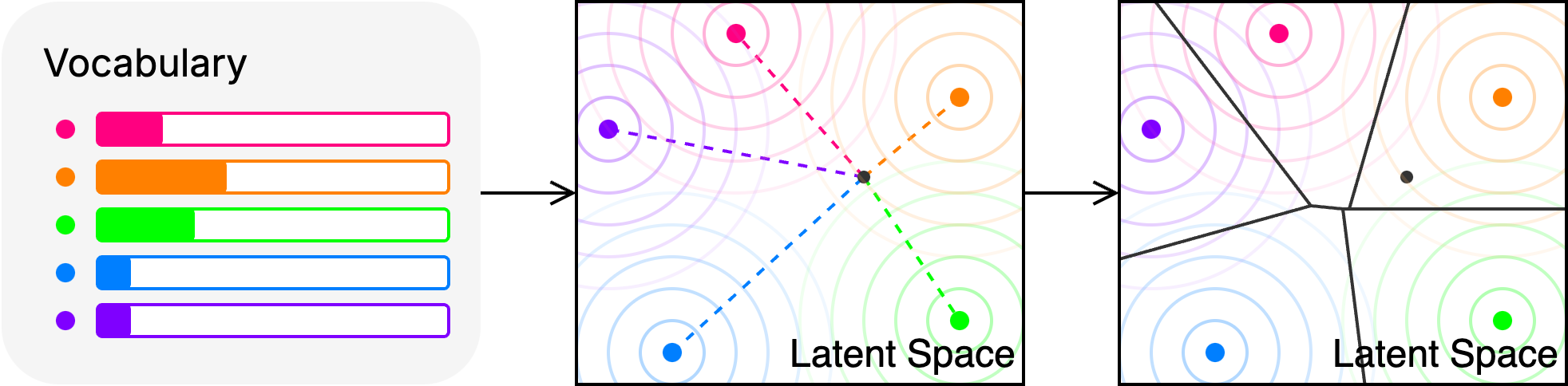}
    \caption{Semantic association of vocabulary and latent space.
    For each color label on the vocabulary (left), there is a color semantic basis in the latent space (middle). The semantics of the dark dot (indicating an arbitrary representation) in the latent space can be quantified as its cosine similarities to semantic bases. The semantics can be computed as probabilities on the vocabulary. When focusing on the nearest semantic basis for a given latent representation, a latent space can be quantified as discrete semantic regions (right).}
    \label{fig:semantics}
\end{figure}

For the recognizable semantic meanings of a given LM, \emph{vocabulary-defined semantics} proposed defining a set of special representations in the latent space to associate with the labels on the vocabulary. It quantifies the semantic property of LM latent space leveraging local isotropy~\citep{Cai2021IsotropyIT}, and benefits parameter optimizations, such as efficient logits computation~\citep{Gu2024VocabularyDefinedSL}.
For each label on the LM vocabulary, there is an associated representation in the latent space, termed as ``semantic basis'', they share the same semantic meaning, as shown in \cref{fig:semantics}.

For a given LM-head matrix, we conduct matrix multiplication to obtain semantic bases in the latent space.
Since the computation direction is from logits to representations, instead of using the LM-head matrix $\mathbb{W}$, we use its pseudoinverse $\mathbb{W}^+$.
If there are $v$ labels in the vocabulary, there will be $v$ unique semantic bases representing all semantic meanings.
At the output side of LM, we multiply each onehot embedding $\vec{e}$ by the pseudoinverse matrix $\mathbb{W}^+$ to obtain the corresponding representation $\vec{s}$. That is, $\vec{s}=\vec{e}\cdot\mathbb{W}^+$.
The computation is equivalent to solving the least squares problem of a system of linear equations.

% \paragraph{Metric for Semantic Logits}
% One the metrics for distance-based similarity, cosine distance is adopted for its direct geometric meaning.
% In rare cases where precise computations on logits are required, instead of the relative magnitude of values (such as the property of ``semantic decomposition''), we suggest using a softmaxed results on the cosine similarities.

\subsection{Hypothesis: Vector Nature of Semantics}

Centered on each semantic basis, there forms a ``semantic field''. The concept of semantic field is similar to the \textit{field} term in physics (such as electric field, then the semantic basis analogies to the electric pole). The semantics of an arbitrary latent representation can be quantified as the overlapping impact of numerous semantic fields, and be further computed as probabilities~\citep{Gu2024VocabularyDefinedSL}.
The process is ``composition of semantics'', where multiple \emph{component semantics} become a \emph{resultant vectors} via vector addition.
Therefore, we assume the overlapping effects of semantic fields support a corresponding reversed operation ``resolution of semantics''. That is, a single \emph{resultant vector} may be resolved into multiple \emph{component vectors} along the directions of semantic bases.

In detail, for a given latent representation $ \vec{r} $, its semantic meaning can be projected to different semantic bases to obtain corresponding ``component semantics'' $ \vec{c_{i}}=\mathtt{proj}(\vec{r}, \vec{s_{i}}) $ (analogy to ``component force'' in a force field). By accumulating the decomposed semantics, we get a ``resultant semantics'' $ \sum\limits_{i=1}^{n} \vec{c_{i}} $ (analogy to ``resultant force'' in a force field).
The equation $ \vec{r} \parallel \sum\limits_{i=1}^{n} \vec{c_{i}} $ stands approximately true. In contrast, when taking a random collection of vectors as semantic bases and obtain $ \vec{c_{i}^\prime}=\mathtt{proj}(\vec{r}, \vec{s_{i}^\prime}) $, the equation $ \vec{r} \perp \sum\limits_{i=1}^{n} \vec{c_{i}^\prime} $ stays true. It is consistent with the property of the latent space that, arbitrary vectors in a high-dimensional space tend to be orthogonal.

\subsection{Empirical Validation}

\begin{figure}[!htb]
    \centering
    \includegraphics[width=\linewidth]{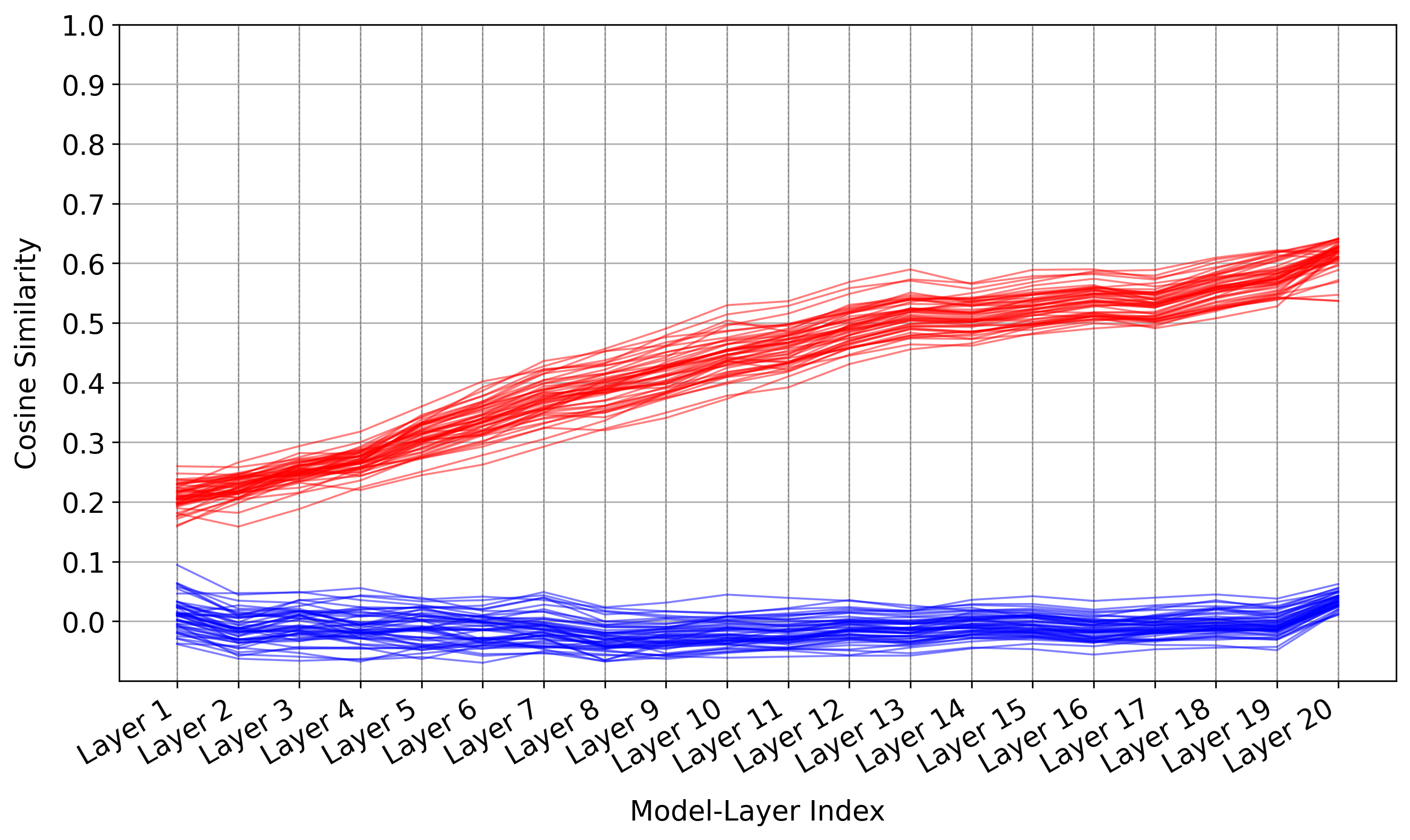}
    \caption{Empirical Validation of Semantics Decomposition (CodeGen).}
    \label{fig:empirical_validation}
\end{figure}

For each label on the LM vocabulary, there is an associated representation in the latent space, termed as ``semantic basis'', they share the same semantic meaning, as shown in \cref{fig:empirical_validation}.

\section{Approach}
\label{sec:approach}

% specify the settings
We introduce the preliminary methods of reducing the information loss when latent representations pass through neighboring modules, as the interaction interface of modules in LM modularization.
Leveraging the semantics isotropy of LM latent space, we discuss two situations of stitching (heterogeneous) modules, with progressive difficulties:
(1) LM vocabularies are the same but LM-head matrices are different;
(2) LM vocabularies are different and LM-heads matrices are different.
In the future, we plan to study the cases of module combination where there are no corresponding module-level vocabularies.
% Meanwhile, we will study the problem of ``module updating'' in the context of LM modularization, or termed as (modular) LM repair.

\subsection{Pairwise Knowledge Fusion}

Pairwise knowledge fusion is a preparatory stage introduced in recent work~\cite{Wan2024FuseChatKF} to enable effective merging of LLMs with diverse architectures or training trajectories. The central idea is to construct intermediate target models by fusing the knowledge of each source model with that of a shared pivot model. These target models are aligned both structurally and semantically with the pivot and can subsequently be merged in the parameter space.

The procedure begins by selecting a single source model as the \textbf{pivot}, typically based on considerations such as architecture compatibility, model size, or performance. Each remaining source model is then paired with the pivot to form a fusion pair. For each pair, the two models are jointly evaluated on a set of prompts to obtain their respective token-level output distributions (i.e., probability matrices over vocabulary tokens). These output distributions are treated as soft representations of the models’ internal knowledge.

\paragraph{Layer-wise Interpolation}
The difficulties in merging heterogeneous LMs are caused by their differences in model structures. For example, even though generative LMs follow a similar hierarchical structure, the amount of model layers differ.
Since the changes between neighboring layers is marginal, we therefore use linear interpolation to estimate the semantics by arbitrary ratios~\cite{Gu2024ASL}.

% PASS
% "Implies successfully transferring and aligning semantics between modules, evoking a smooth and successful transition".
% PASS – Preserving-Aligned Symbolic-Semantics
% Dynamic and ongoing: This version emphasizes the active, continuous process of maintaining alignment and semantics.
% Proactive tone: It suggests your approach is consistently working to safeguard semantics during transitions.
% Best for methods emphasizing ongoing processes
% If your approach emphasizes an ongoing and dynamic process, go with "Preserving Aligned Symbolic Semantics."

% TODO say somewhere what we will do as next steps, and refine all other contents...
%  such as the last parts of analysis and approach, as well as the introduction.
%  check all other details...
Our current approach is merely for the case where two neighboring modules share the same vocabulary (but LM-head matrices are different), once its usefulness is verified, we will go further to study the case where the vocabularies of neighboring modules are different.
On the approach described as follows, we first introduce a useful geometric property of the representations in LM latent space, and then, summarize the operations to guarantee information lossless when latent representations pass through neighboring modules.

\subsection{Semantics-Preservative Transformation}

We term the transformation on latent representations that remain the same probabilities in neighboring modules as ``semantics-preservative transformation''.
Assume a given latent representation $\vec{r}$ passes from module $M_x$ to module $M_y$, for the representation $\vec{r_x}$ of $M_x$, we can directly solve the semantics-preservative representation $\vec{r_y}$ of $M_y$ leveraging semantics decomposition.

% no need to mention too many details
We first study the case where neighboring modules share the same vocabularies, and our approach is as follows (once it is done, we go further to study the case where the vocabularies are different):
\begin{enumerate}
    \item compute the semantic bases of neighboring modules, note as $\vec{s_x}$ and $\vec{s_y}$;
    \item compute the semantic probabilities based on the cosine similarities between representation $\vec{r_x}$ and corresponding semantic bases $\vec{s_x}$;
    \item estimate the semantics-preservative representation $\vec{r_y}$ using a weighted linear combination on semantic bases $\vec{s_y}$, respecting the computed probabilities;
    \item the magnitude of $\vec{r_y}$ may require additional calibrations (to be confirmed by further experiments).
\end{enumerate}

\subsection{Semantic Alignment}

% TODO the emperial evidence for semantic field... (in the appendix)

Due to differences in tokenizer vocabularies and response sequences across models, direct comparison of these distributions is not straightforward. To address this, we realizes \textbf{semantic alignment} to solve two concerns:

% Multiple Tokens as One Token
% One Token as Multiple Tokens

\paragraph{Sequence Segmentation}
Models often produce token sequences of different lengths and granularities in response to the same input. Sequence alignment reconciles these differences by mapping tokens from the source model to those of the pivot. This is typically accomplished using dynamic programming to minimize edit distance, supporting many-to-one, one-to-many, and one-to-one mappings. The result is an aligned sequence space in which model outputs can be compared positionally.

\paragraph{Vocabulary Mapping}
Even when sequence positions are aligned, the vocabularies used by different tokenizers may not match. Vocabulary alignment maps tokens from the source model's vocabulary to those of the pivot model. Early approaches used exact token matching, which limited coverage. Later methods relaxed this by allowing fuzzy matching based on edit distance. More recent techniques utilize token mapping statistics—aggregated from aligned sequences—to guide the alignment, enabling more reliable handling of one-to-many or many-to-one token correspondences.

\begin{figure}[!htb]
    \centering
    \includegraphics[width=\linewidth]{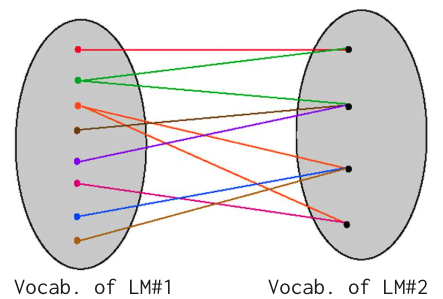}
    \caption{Semantic alignment solving the concerns of sequence segmentation and vocabulary mapping.}
    \label{fig:mapping}
\end{figure}

Once both alignment stages are complete, the output distributions from the source and pivot models are fused using a predefined strategy such as minimum cross-entropy. The resulting distributions serve as soft targets for constructing a target model that integrates knowledge from both source models. These target models, all aligned to the pivot’s structure, can then be merged efficiently in the parameter space.

While effective, this fusion strategy depends on model outputs, alignment heuristics, and a dataset of prompts. In this work, we aim to eliminate these dependencies by directly merging model parameters through a semantic-based analysis, without requiring any alignment or data sampling.

\section{Analysis}

% objective
The requirement of the development protocol can be clarified from the perspective of information, that is, making the intermediate information of a leading module be fully retained and delivered to subsequent modules, until the last module.
Therefore, the requirement includes two problems: (1) \ul{what is the essential information in passing between modules}, and (2) \ul{how to realize information lossless in passing between modules}.
We answer the problems from a semantic-based perspective, where the information is seen as the quantification of the semantic property of data (tokens, representations, etc).
We focus on the interactions in the form of latent representations, which are more informative than using human-readable signals (such as tokens).

% answer to problem 1
% % definition
% In the context of modular language models, we term \emph{global vocabulary} as the vocabulary used in model-level inputs and outputs (corresponding to the input-side embedding matrix, and the output-side LM-head matrix); similarly, we term \emph{local vocabulary} as the vocabulary used in module-level inputs and outputs, which varies by module.
For each token, both the initial information at the model-level input-side and the resulting information at the model-level output-side is the probabilities on the model-level vocabulary.
Therefore, \ul{the essential information passing between modules is the probabilities on the model-level vocabulary}.
In contrast, when the interaction medium is human-readable symbols (such as tokens), the information is discrete labels, namely the hardmax results of probabilities.
% Assume a given sequence of $n$ tokens, $t_1, t_2, ..., t_{n}$
Assume the size of model-level vocabulary is $v$, the probabilities will be: $p_1, p_2, ..., p_{v}$. The modules shall take a superset of the model-level vocabulary or share the same vocabulary.

% answer to problem 2
Between modules, latent representations are the intermediate results and the medium of computation.
% They implicitly contains the information of probabilities.
Assume the dimension of a latent space is $d$, the representation will be: $r_1$, $r_2$, ..., $r_d$.
Compared with the probabilities on vocabulary, the representations in latent space have a much smaller size (in language models, $d$ is much smaller than $v$).
It implies that LM-head matrix plays the role of eliminating the information gap between representations and probabilities.
When neighboring modules share the same LM-head, information remains the same when representations pass through modules. However, if LM-heads of neighboring modules are different, the representations shall transform based on their difference to reduce the information loss.
That is, \ul{latent representations shall transform respecting the LM-heads of modules to realize information lossless}.

% ...
% When both LM vocabularies and LM-heads are not available, the semantics in latent space cannot stand local isotropy as well. However, the situation will be simple, since the information will be discrete, instead of continuous.
% since it will be tokenized input, just like the onehot embeddings, just one-hot information...

% the information depends on the for the same representation, the probabilities computed with dictionaries of neighboring modules remain same

% To guarantee the information loss will not happen in the computation using latent representation, we can introduce a ``dictionary'' to each module to provide the additional information. Here, ``dictionary'' means a technique of complementing the missing information, returning a representation (a vector of size $d$) for the given probabilities (a vector of size $v$).
% Besides, ``dictionaries'' of neighboring modules can examine that the representations (in the common latent space) are information-lossless when passing between modules.

% name our approach of aligning feature spaces...
% The essence of xxx lies in aligning the feature spaces of heterogeneous modules.

% the approach is composed of different practices, most practices are only in need when the corresponding condition are noy fulfilled (each practice is to realize the corresponding condition, so when the condition is already there, the practice will not in need)
% Depending the semantic property of LM latent space, we may need to take different practices into our approach.

\section{Related Work}
\label{sec:related_work}

\emph{LM fusion} indicates a wide range of model collaboration strategies, and gained significant attention due to the potential to integrate the strengths of multiple individual models.
In this paper, we focus on a concrete topic, that is, model merging. Compared to other collaboration paradigms of LM fusion, model merging directly operates at the weight level to synthesize a single performant model from sources models.

\paragraph{Model Merging}
Early works on merging focused on simple arithmetic combinations of model parameters. \citet{Wortsman2022ModelSoup} introduced \emph{Model Soup}, which demonstrated that averaging the weights of multiple fine-tuned models can yield better generalization. \citet{Matena2021Fisher} extended this idea with \emph{Fisher merging}, weighting parameters by their estimated importance using the Fisher Information Matrix. More recent efforts focus on resolving parameter conflicts that arise when merging models trained on different tasks. \citet{Ilharco2023TaskArithmetic} proposed \emph{task arithmetic}, leveraging the difference vectors between pre-trained and fine-tuned models to represent task semantics. Building on this, \citet{Yadav2023TIES} introduced \emph{TIES-Merging}, which mitigates sign conflicts and redundant updates, while \citet{Yang2024AdaMerging} proposed adaptive, layer-wise merging guided by test-time objectives. \citet{Jin2023RegMean} formulated merging as a regression problem, learning optimal merge weights in closed form. While effective, these techniques typically require labeled or unlabeled data to compute merge coefficients, making them resource-intensive and susceptible to bias from sampling.

\paragraph{LM Fusion}
The concept of LM fusion is complex, since it is not about technical, but an application term. It emphasizes using capabilities from different models, such as LM ensemble, LM routing, and LM mixing.
\emph{Model ensemble} methods combine model outputs rather than weights. In inference, all ensembled models will be activated. Though widely used, they introduce high inference costs and do not yield a unified model.
\emph{Model routing} strategies coordinate multiple LLMs through an individual routing module, often requiring flexible and proper decisions on which model to activate and which not to. The retraining is optional, depending on the actual practice.
\emph{Model mixing} goes further by structurally integrating components (e.g., layers or experts) from different models, frequently resulting in larger models that require retraining. These strategies differ fundamentally from merging in that they operate at runtime or alter the model architecture, while merging seeks a compact, unified model in parameter space.

In this work, we focus exclusively on parameter-level model merging due to its scalability, inference efficiency, and growing importance for model reuse and multi-task generalization.

\section{Conclusion}
\label{sec:conclusion}

In this paper, we have proposed the concept of semantic transition. By defining semantic trace and semantic route as factual and virtual transitions, we explain LM finetuning as the process of letting the factual one steer to the virtual one in latent space.
% The deviation between them in each layer will be reduced so the last-layer representation will be close to the ground truth.
Further, we propose semantic-based layer-freezing to accelerate LM finetuning, by finding the layer with the least deviation and freeze the deeper layers.
Based on our results, semantic-based layer-freezing provides better performance than the state-of-the-art as well as the common practices.
% It is also noticeable that layer-freezing tends to have better performance than full-layer finetuning.
% We also introduce budget plans to control the cost-benefit tradeoff.
Moreover, our work explores the effects of budget plans on the cost-benefit tradeoff for layer-freezing.
% By following the arithmetic growth and depth-first infilling, the finetuning has the least computation cost while realizing the best performance.
% 
In return, the effectiveness of our lay-finetuning approach validates the usefulness of semantic transition.

\bibliography{custom}

% \appendix

% \input{sections/appendix}
% \input{sections/discussion}

\end{document}